\UseRawInputEncoding
\documentclass[letterpaper, 10 pt, conference]{ieeeconf}  % Comment this line out if you need a4paper

\IEEEoverridecommandlockouts                              % This command is only needed if 
\usepackage{longtable}% for long tables
\usepackage{booktabs}
\usepackage{algorithm}
\usepackage{bbm}
\usepackage{xcolor}
\usepackage{subfigure}
\usepackage{amsfonts}
\usepackage{amsmath,amssymb,amsfonts}
\usepackage{algorithmic}
\usepackage{graphicx}
\usepackage{textcomp}
\usepackage{xcolor}
\usepackage{subfigure}
\usepackage{makecell}
\usepackage{cite}
\usepackage{graphicx}
\usepackage{amsmath}
\usepackage{amssymb}
\usepackage{booktabs}
\usepackage{multicol}
\usepackage{multirow}
\usepackage{bm}
\usepackage{comment}
\usepackage{booktabs}
\usepackage{array}
\usepackage{hyperref}
\usepackage{cleveref} 
\usepackage{booktabs}
\usepackage{array} % 提供了扩展的列定义选项，如 m for vertical centering
\usepackage{siunitx} % 使用siunitx来对齐数字
\Crefformat{figure}{#2Fig.~#1#3}
\Crefmultiformat{figure}{Figs.~#2#1#3}{ and~#2#1#3}{, #2#1#3}{ and~#2#1#3}
\newcommand{\vect}[1]{\boldsymbol{#1}}
\newcommand{\mat}[1]{\mathbf{#1}}

\begin{document}
\title{\LARGE \bf
%Motion-Aware Dynamic Latent Diffusion Model for Tele-operated Robot Video Debluring
MotionGS-SLAM: Event-Modulated Gaussian Splatting for Motion-Blur Robust SLAM
{  
}}
\author{Zhiqiang HU, Shouren HUANG and Masatoshi ISHIKAWA% <-this % stops a space
\thanks{}% <-this % stops a space
\thanks{The authors are with the Research Institute for Science \& Technology, Tokyo University of Science
        {\tt\small \{zhiqiang.hu, huang, ishikawa\}@ishikawa-vision.org}}%
}

\maketitle
\thispagestyle{empty}
\pagestyle{empty}

\begin{abstract}

Current Vision-based SLAM systems fail catastrophically when motion blur corrupts the visual input, as they attempt the ill-posed inverse problem of recovering sharp content from degraded observations. We present \textbf{MotionGS-SLAM}, which fundamentally reimagines motion blur handling through a paradigm shift: rather than removing blur artifacts, we reformulate the challenge as a well-constrained forward problem that generatively models blur formation within the rendering pipeline. By leveraging event cameras' microsecond temporal resolution and immunity to motion blur, we introduce a novel \textbf{event-modulated Gaussian kernel} that dynamically adapts each Gaussian's rasterization based on precise motion cues. Our dual-modulation mechanism transforms 2D Gaussian projections from isotropic dots into anisotropic, motion-aligned elliptical brush strokes (spatial modulation) while adaptively varying exposure integral sampling density based on local velocity (temporal modulation). This physics-based approach enables joint optimization of intra-exposure camera trajectories and 3D scene geometry through blur-aware photometric and event-based constraints. Extensive experiments demonstrate significant improvements over state-of-the-art methods in trajectory accuracy and map quality under severe high-motion conditions.
\end{abstract}

%%%%%%%%% BODY TEXT
\section{Introduction}
\label{sec:intro}

Simultaneous Localization and Mapping (SLAM) is a cornerstone of robot autonomous systems, empowering robots to navigate and interact with unknown environments~\cite{orbslam3, droidslam}. While classical visual SLAM systems excel at localization, they often produce sparse geometric maps. The recent introduction of 3D Gaussian Splatting (3DGS)~\cite{gs} into the SLAM pipeline has marked a significant leap forward, enabling the creation of dense, photorealistic scene representations in real-time~\cite{gsslam, splatam, monogs}. These GS-SLAM systems hold immense promise for applications like augmented reality and digital twinning.

However, the impressive performance of current GS-SLAM methods is predicated on a critical assumption: the availability of high-quality, sharp input images. This assumption frequently breaks down in real-world robotic applications, particularly during aggressive motion or in low-light environments that necessitate long exposures as shown in Fig. 1. To compensate for dim lighting, cameras rely on long exposure times, which inevitably leads to severe motion blur. This blur corrupts the high-frequency details essential for feature matching, resulting in degraded camera tracking and, consequently, a corrupted 3D Gaussian map filled with artifacts. While some approaches have attempted to tackle this by integrating deblurring modules like~\cite{deblurnerf}, they often struggle with severe, persistent blur as they attempt to \textit{invert} a highly degraded signal rather than modeling the underlying physical process.
\begin{figure}[t]
    \centering
    \includegraphics[width=0.88\linewidth]{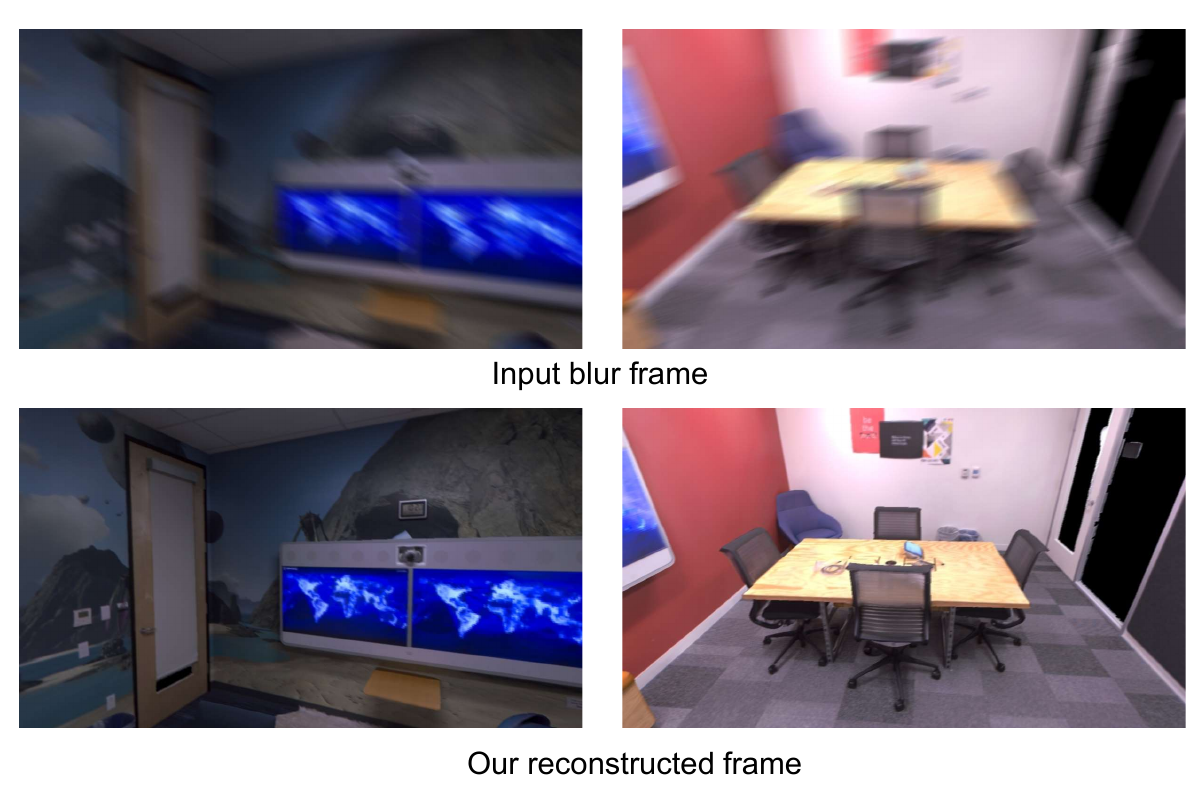}
    \caption{\textbf{MotionGS-SLAM in action.} Existing Vision-based SLAM systems fail in challenging high-motion scenarios due to severe motion blur in the input frames (top row). Our method leverages the co-located event stream to explicitly model the physical process of blur formation. This allows MotionGS-SLAM to robustly track the camera and reconstruct sharp, high-fidelity scenes (bottom row) directly from the degraded imagery, where other methods would fail.}
    \label{fig:teaser}
\end{figure}
\begin{figure*}[t]
    \centering
    \includegraphics[width=\textwidth]{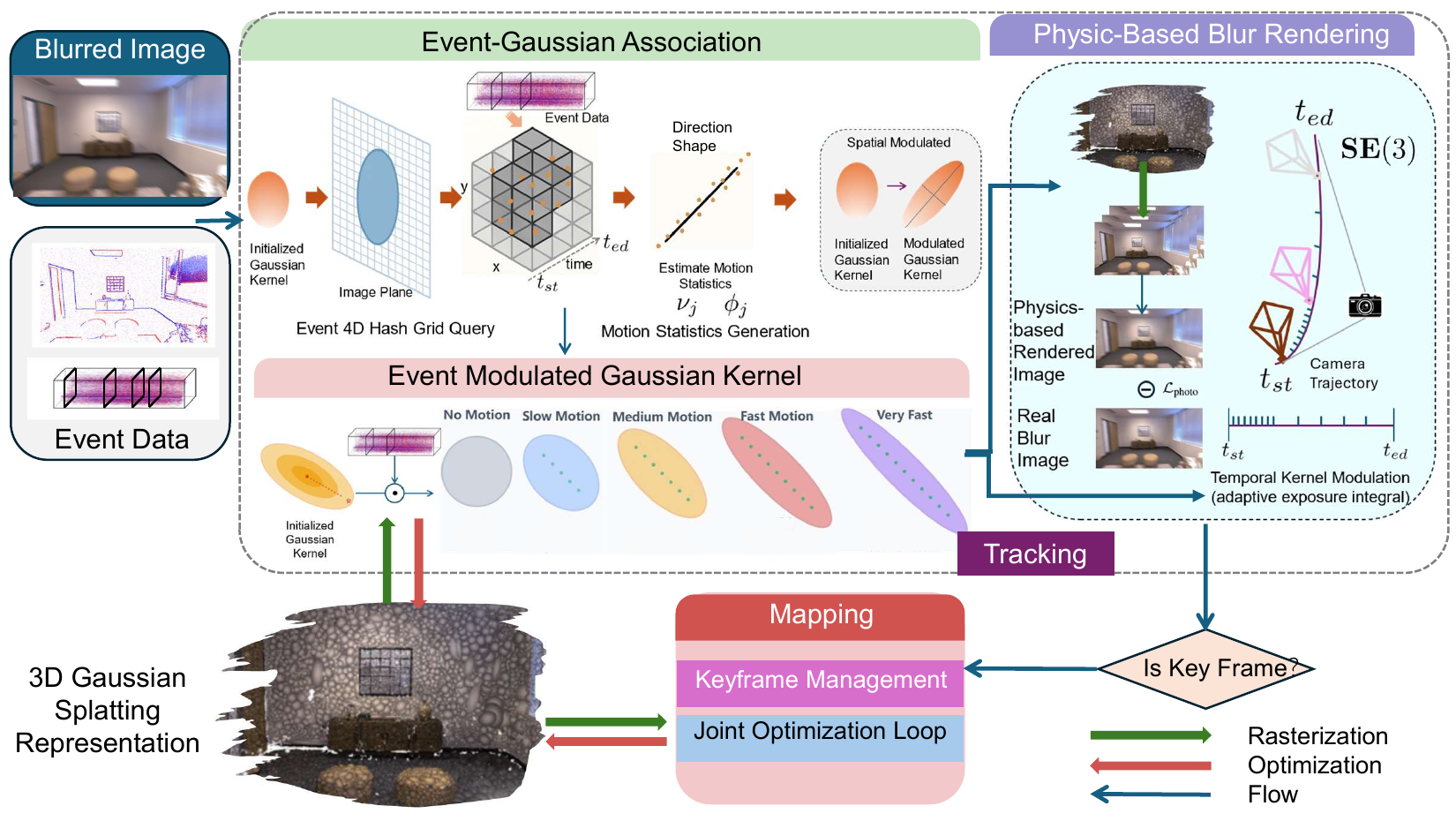} % Replace with your image file name
    \caption{\textbf{Overview of the MotionGS-SLAM pipeline.} Our system takes a blurry RGB image and a stream of events as input. The core of our method is the \textbf{Event-Modulated Gaussian Kernel}, which leverages events to physically model motion blur. \textbf{(1) Event-Gaussian Association:} For each projected Gaussian, we use a 4D spatio-temporal hash grid to efficiently query associated events. From these, we compute local motion statistics (velocity $\nu_j$, direction $\phi_j$). \textbf{(2) Kernel Modulation:} These statistics drive our novel dual-modulation mechanism, anisotropically scaling the projected 2D Gaussian's shape (spatial modulation) to replicate the motion streak. \textbf{(3) Physics-Based Rendering:} Our renderer uses the modulated kernels and an interpolated camera trajectory ($T_{st}$, $T_{ed}$) to synthesize a blurred image that matches the real observation. This entire differentiable process is used in two threads: a real-time \textbf{Tracking} front-end that optimizes only the camera trajectory, and a \textbf{Mapping} back-end that performs joint optimization over a keyframe window to refine both the 3D Gaussian map and poses.}
    \label{fig:pipeline}
\end{figure*}

The fundamental insight of this work is a paradigm shift in how we approach motion blur in visual SLAM. Rather than treating motion blur as an undesirable artifact to be removed through deconvolution, we embrace it as the natural physical consequence of camera motion during exposure and model it generatively. Our key observation is that traditional deblurring methods attempt to solve a severely ill-posed inverse problem recovering sharp content from degraded observations where critical high-frequency information has been irreversibly lost. In contrast, we reformulate this challenge as a well-constrained forward problem: leveraging the high-temporal-resolution motion cues from event cameras, we physically simulate the blur formation process within our differentiable renderer, actively ``painting'' blur that matches the observed degraded images.

To realize this vision, we introduce MotionGS-SLAM. Our system models the intra-exposure camera trajectory as a continuous path in $\mathbf{SE}(3)$ space. Given the typically brief exposure intervals, we parameterize each trajectory segment using only its boundary poses the camera configurations at the exposure's onset and conclusion. During tracking, we optimize these pose pairs by enforcing photometric consistency between the observed blurry frames and our physically-rendered blur simulations. This joint constraint of event-guided motion estimation and blur-aware photometric alignment substantially enhances tracking precision. In the mapping phase, we perform joint optimization over both the trajectories of strategically selected keyframes and the 3D Gaussian scene representation. This optimization minimizes a combined objective of photometric reconstruction error and event-based geometric constraints, resulting in sharp, geometrically accurate maps despite severely degraded input imagery.

The technical cornerstone of our approach is an  \textbf{event-modulated Gaussian kernel} that dynamically adapts how each 3D Gaussian is rasterized onto the image plane. This adaptation operates through a dual-modulation mechanism: \textbf{Spatial Modulation:} Guided by local motion statistics extracted from the event stream, we transform each Gaussian's 2D projection from an isotropic circular ``dot'' into an anisotropic elliptical ``brush stroke.'' The ellipse's orientation aligns with the detected motion direction while its eccentricity scales with motion magnitude, geometrically replicating the characteristic streak patterns of motion blur.
\textbf{Temporal Modulation:} We adaptively adjust the temporal sampling density of the exposure integral based on the scene's aggregate motion. Frames with higher average motion velocity, computed across all visible Gaussians, are rendered with a greater number of discrete camera poses.This strategy accurately models the cumulative photometric effect while maintaining computational efficiency.

Our main contributions are summarized as follows:
\begin{itemize}
    \item We propose \textbf{MotionGS-SLAM}, that tackles motion blur from a physics-based perspective. Instead of treating blur as an artifact to be removed, we model its physical formation process within the renderer, guided by an event camera.
    \item We introduce an \textbf{event-modulated Gaussian kernel} featuring a dual-modulation strategy. It dynamically adapts the spatial shape and temporal sampling density of projected Gaussians based on local motion statistics derived efficiently from events via a 4D hash grid.
    \item We demonstrate that our system, which requires no depth sensor, significantly outperforms state-of-the-art visual and GS-SLAM methods in both tracking accuracy and reconstruction quality on challenging synthetic and real-world sequences with severe motion blur.
\end{itemize}

%-------------------------------------------------------------------------
\section{Related Works}
\subsection{Gaussian Splatting based SLAM}

The introduction of 3D Gaussian Splatting~\cite{gs} has recently enabled a new class of SLAM systems that create dense, photorealistic maps in real time. Pioneering works like GS-SLAM~\cite{gsslam}, SplaTAM~\cite{splatam}, MonoGS~\cite{monogs}, Sgs-slam ~\cite{li2024sgs}, Large spatial model~\cite{fan2024large} and Compact 3d gaussian splatting~\cite{deng2024compact} successfully integrated 3DGS into a unified tracking and mapping framework, demonstrating remarkable results on high-quality video sequences. These systems established the viability of using an explicit, differentiable representation for both localization and scene reconstruction.

However, the core assumption of sharp, well-exposed images makes these methods brittle. Recognizing this, subsequent works have attempted to address the challenge of motion blur. While approaches like I\textsuperscript{2}-SLAM~\cite{i2slam} model the camera's imaging process, these image-only methods are fundamentally constrained. They attempt to recover lost information from an already corrupted signal, an approach that fails when severe motion makes the blur too ambiguous to reliably invert. Our work differs by leveraging a secondary, high-frequency modality to directly inform the rendering of motion, rather than relying solely on the degraded image.

\subsection{Event-based 3D Gaussian Splatting/NeRF}

The asynchronous nature and inherent immunity to motion blur make event cameras particularly advantageous for three-dimensional scene reconstruction.  A recent surge of research has focused on combining event streams with NeRF/3DGS. E2NeRF~\cite{qi2023e2nerf} integrates both blurred image supervision and event-based constraints to recover high-fidelity NeRF representations from severely degraded inputs. Building upon this, Ev-DeblurNeRF~\cite{cannici2024mitigating} introduces a learned event-to-intensity mapping that effectively suppresses event noise, leading to enhanced reconstruction fidelity. Meanwhile, the method in~\cite{eventnerfslam} demonstrates the feasibility of 
event-guided implicit neural SLAM by fusing event streams with RGB-D 
inputs, achieving robust tracking under motion blur and lighting 
variation through a differentiable CRF rendering technique. Methods like EvaGaussians~\cite{yu2024evagaussians} and Event3DGS~\cite{xiong2024event3dgs, han2024event} have shown that by leveraging event data, it is possible to reconstruct crisp, detailed 3D Gaussian scenes from a set of blurry images. These works successfully demonstrate the power of events for high-fidelity reconstruction.
However, these are primarily offline reconstruction pipelines that require pre-computed camera poses (e.g., via SfM), rendering them incapable of online SLAM. Our work bridges this critical gap by integrating our event-guided generative rendering directly into the front-end tracking loop. This enables a complete and robust online GS-SLAM system that succeeds where prior methods fail.
%%%%%%%%%%%%%%% ICRA2025 METHOD (MODIFIED) START %%%%%%%%%%%%%%%

\section{Method}
\label{sec:method}
 The core architecture of our method is shown in Fig. 2.

\subsection{Motion Blur Image Formation Model}
\label{sec:phys_render}

We represent the scene as a set of 3D Gaussians $\mathcal{G} = \{g_i\}$, where each Gaussian is defined by a mean $\vect{\mu}_i \in \mathbb{R}^3$, covariance $\mat{\Sigma}_i \in \mathbb{R}^{3\times3}$, color $\vect{c}_i$, and opacity $o_i$.
In a standard static rendering, each 3D Gaussian is projected onto the 2D image plane given a single camera pose $\mat{T}$, and the final color $\mat{C}(\vect{u})$ at a pixel $\vect{u}$ is computed via $\alpha$-blending.

However, motion blur is a physical process of light integration over a non-zero exposure interval $\Delta t_m$, during which the camera is moving. We model the camera's continuous trajectory $\mat{T}(t)$ for $t \in [t_{st}, t_{ed}]$ by interpolating a start pose $\mat{T}_{st}$ and an end pose $\mat{T}_{ed}$. The resulting blurred image $\bar{\mat{C}}_m$ is the temporal integral of infinitesimally short exposures along this path. This is approximated by discretely summing $N$ latent sharp images:
\begin{equation}
    \bar{\mat{C}}_m(\vect{u}) \approx \frac{1}{N} \sum_{k=1}^{N} \mat{C}(\mat{T}(t_k), \vect{u}).
    \label{eq:exp_int}
\end{equation}
where $\mat{C}(\mat{T}(t_k), \vect{u})$ is the sharp image rendered at an intermediate pose $\mat{T}(t_k)$.

\noindent\textbf{Camera Motion Trajectory Modeling}
\label{sec:trajectory}
 We parameterize the camera motion using two poses: the exposure start $\mat{T}_{st} \in \mathbf{SE}(3)$ and end $\mat{T}_{ed} \in \mathbf{SE}(3)$. The pose at any time $t \in [0, 1]$ during exposure is:
\begin{equation}
    \mat{T}(t) = \mat{T}_{st} \cdot \exp(t \cdot \log(\mat{T}_{st}^{-1} \cdot \mat{T}_{ed})),
    \label{eq:trajectory}
\end{equation}
where $\exp$ and $\log$ are the exponential and logarithm maps on $\mathbf{SE}(3)$.

\subsection{Motion Blur Aware Tracker}
\noindent\textbf{Real-Time Event-Gaussian Association}
\label{sec:assoc}

To provide each Gaussian with its local motion context, we must efficiently associate it with the relevant subset of events. A brute-force search is computationally infeasible for real-time operation.

\textbf{4D Hash Grid.} We employ a multi-resolution hash grid to index events in a 4D spatio-temporal domain. Each event $e = (x, y, t, p)$ is characterized by its spatial coordinates $(x, y) \in \mathbb{R}^2$, timestamp $t \in \mathbb{R}$, and polarity $p \in \{-1, +1\}$ indicating brightness decrease or increase, respectively.

\textbf{Query Process.} 
For each visible 3D Gaussian $g_j$, we project it to the 2D image plane, obtaining its mean $\hat{\vect{\mu}}_j$ and covariance $\hat{\mat{\Sigma}}_j$. 
We then query the hash grid for all events within the $3\sigma$ axis-aligned bounding box of the 2D Gaussian and within the camera's exposure window $[t_{st}, t_{ed}]$. 
We refine this candidate set by retaining only the events $\mathcal{E}_j$ that satisfy the Mahalanobis distance criterion:
\begin{equation}
    (\vect{p} - \hat{\vect{\mu}}_j)^\top \hat{\mat{\Sigma}}_j^{-1} (\vect{p} - \hat{\vect{\mu}}_j) \le \tau,
    \label{eq:mahal}
\end{equation}
where $\tau$ is a tunable threshold (typically $\tau\!\approx\!9$, corresponding to the 99\% confidence contour of a 2D Gaussian). 
This hash-based approach reduces the association complexity from $O(N_\text{gauss} \times N_\text{event})$ to an efficient $O(N_\text{gauss} \times K)$, where $N_\text{gauss}$ is the number of visible Gaussians in the current frame, $N_\text{event}$ is the number of events in $[t_{st},t_{ed}]$, and $K$ is the small average events returned per Gaussian by the grid query.
%%%%%%%%%%

\textbf{Motion Statistics.}
For each Gaussian \(g_j\), we estimate its local motion statistics from the associated event set \(\mathcal{E}_j\) on the image plane.
Specifically, we first compute a polarity-weighted average displacement of events relative to the Gaussian center:

\begin{equation}
\begin{aligned}
    \vect{v}_j &=
    \frac{\sum_{e_i \in \mathcal{E}_j} p_i\,(\mathbf{x}_i - \hat{\vect{\mu}}_j)}
         {\sum_{e_i \in \mathcal{E}_j} |p_i| + \varepsilon},
    \\[4pt]
    \nu_j &= \|\vect{v}_j\|,\qquad 
    \phi_j = \operatorname{atan2}(v_y, v_x),
\end{aligned}
\label{eq:stats}
\end{equation}
where \(p_i \in \{-1,+1\}\) is the polarity and 
\(\mathbf{x}_i \in \mathbb{R}^2\) is the 2D spatial coordinate 
of event \(e_i\), \(\hat{\vect{\mu}}_j\) is the projected 2D mean 
of Gaussian \(g_j\), and \(\varepsilon\) is a small constant to 
avoid division by zero.
The resulting vector \(\vect{v}_j\) represents the dominant local motion flow: events with opposite polarities moving in the same direction reinforce each other, while inconsistent noise cancels out.
We use its magnitude \(\nu_j\) as the estimated motion speed and its angle \(\phi_j\) as the motion direction.

%%%%%%%
\begin{figure*}[t]
    \centering
    \includegraphics[width=0.88\linewidth]{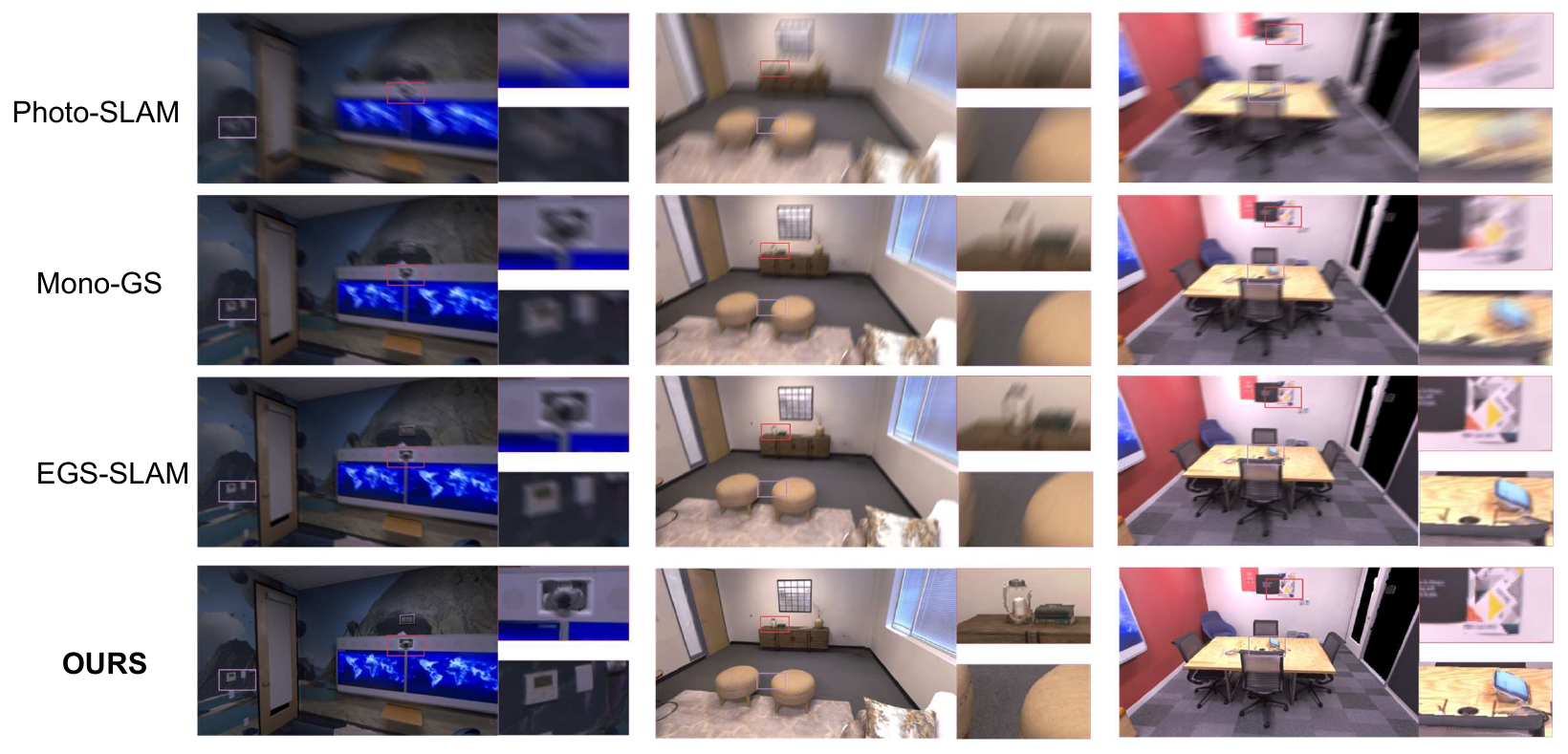} 
    \caption{\textbf{Qualitative comparison of reconstructed scenes under severe motion blur conditions.} The top three rows show results from state-of-the-art baselines: Photo-SLAM~\cite{huang2024photo}, Mono-GS~\cite{monogs}, and EGS-SLAM~\cite{chen2025egs}. Photo-SLAM and Mono-GS struggle significantly, producing highly blurry and distorted reconstructions with numerous artifacts and "ghosting" effects. EGS-SLAM, while benefiting from event data, still exhibits noticeable blur and less accurate details, particularly in fine textures (e.g., the television screen, object edges). In stark contrast, the bottom row showcases results from \textbf{MotionGS-SLAM (Ours)}. Our method consistently reconstructs sharp, high-fidelity scenes with clear details and accurate geometry.}
    \label{fig:comparison_qualitative}
    \vspace{-0.8em}
\end{figure*}
%%%%%%%%%%%%%
%%%%%%%%%%%%%
\noindent\textbf{Event-Modulated Gaussian Kernel}
\label{sec:mod_kernel}

The computed motion statistics $(\phi_j, \nu_j)$ drive our dual-modulation kernel, which adapts both the spatial and temporal aspects of the rendering process for each Gaussian. Our approach separates the \emph{true 3D geometry} from \emph{motion blur effects} through a two-component covariance design, ensuring the core 3D map remains geometrically accurate while enabling physics-based blur synthesis.

\paragraph{Spatial Kernel Modulation.}
To model the spatial characteristics of motion blur, we construct a motion-aligned prior covariance from event measurements:
\begin{equation}
\begin{aligned}
    \mat{\Sigma}^{2D}_{j,\text{prior}}
    &= \mat{R}(\phi_j)\;
       \mathrm{diag}\!\big(\sigma_\perp^2,\ \kappa_j^2 \sigma_\perp^2\big)\;
       \mat{R}(\phi_j)^\top,
    \\[4pt]
    \kappa_j &= 1 + \beta\nu_j,
\end{aligned}
\label{eq:sigma_prior}
\end{equation}
where $\mat{R}(\cdot)$ is the 2D rotation matrix, $\sigma_\perp^2$ is a base variance, and $\kappa_j$ is a velocity-dependent stretch factor. This formulation creates an anisotropic Gaussian aligned with the motion direction $\phi_j$ and elongated proportionally to the velocity $\nu_j$, replicating the characteristic streak pattern of motion blur.

% Eq. (6) - Spatial Kernel Modulation 段落中
We enforce this motion-consistent shape through a regularization loss:
\begin{equation}
    \mathcal{L}_{\text{shape}} = 
    \frac{1}{|\mathcal{G}_{\text{vis}}|}
    \sum_{j \in \mathcal{G}_{\text{vis}}} 
    \big\|\hat{\mat{\Sigma}}^{2D}_j - \mat{\Sigma}^{2D}_{j,\text{prior}}\big\|_F^2,
    \label{eq:lshape}
\end{equation}
where $\hat{\mat{\Sigma}}^{2D}_j$ is the projected 2D covariance of the 
optimized 3D Gaussian, and $\mathcal{G}_{\text{vis}}$ is the set of 
Gaussians visible in the current frame.

To maintain numerical stability while separating geometry from blur, we decompose the 3D covariance as:
\begin{equation}
\begin{aligned}
    \mat{\Sigma}_j &= \mat{\Sigma}_{\text{base},j} + \Delta\mat{\Sigma}_j, \\
    \mat{\Sigma}_{\text{base},j} &= \mat{R}(q_j) \operatorname{diag}(e^{\ell_{x,j}}, e^{\ell_{y,j}}, e^{\ell_{z,j}}) \mat{R}(q_j)^\top,
\end{aligned}
\label{eq:sig3d}
\end{equation}
where $\mat{\Sigma}_{\text{base},j}$ encodes the static 3D geometry with guaranteed positive-definiteness through exponential parameterization, and $\Delta\mat{\Sigma}_j$ represents the motion-induced deformation bounded via activation functions. This decomposition ensures that motion blur modeling does not corrupt the underlying geometric representation the optimizer learns $\Delta\mat{\Sigma}_j$ indirectly through the shape prior (Eq.~\eqref{eq:lshape}) while maintaining photometric consistency.

\paragraph{Temporal Kernel Modulation (Adaptive Sampling).}
Motion blur arises from integrating light over the exposure interval, during which both the camera and scene features move continuously. To accurately model this process, we adaptively sample the camera trajectory based on the aggregate motion characteristics of all visible Gaussians. 

For each frame $m$, we determine the trajectory sampling density by considering the motion statistics across all Gaussians:
\begin{equation}
\bar{\nu}_m = \frac{1}{|\mathcal{G}_{\text{vis}}|} \sum_{j \in \mathcal{G}_{\text{vis}}} \nu_j,
\label{eq:avg_velocity}
\end{equation}
where $\mathcal{G}_{\text{vis}}$ denotes the set of visible Gaussians and $\bar{\nu}_m$ represents the scene's average motion magnitude.

This aggregate motion drives our adaptive sampling strategy:
\begin{equation}
N_m = \mathrm{clamp}\!\left( 
    \mathrm{round}\!\left(n_0 \cdot (1 + \alpha \cdot \bar{\nu}_m \cdot \Delta t_m)\right),
    N_{\min}, N_{\max}
\right),
\label{eq:adaptive_samples}
\end{equation}
where $n_0$ is the base sampling rate, $\alpha$ scales the motion response, $\Delta t_m$ is the exposure duration, and the result is clamped to $[N_{\min}, N_{\max}]$ for computational efficiency.

As illustrated in the Fig. 2, this mechanism creates a motion-aware sampling of the camera trajectory: rapid scene motion (high $\bar{\nu}_m$) triggers denser sampling to capture the complex blur formation, while slow or static scenes require fewer samples. The camera poses are then uniformly distributed along the SE(3) geodesic between $\mat{T}_{st}$ and $\mat{T}_{ed}$:
\begin{equation}
t_k = t_{st} + k \cdot \frac{\Delta t_m}{N_m - 1}, \quad k \in \{0, 1, ..., N_m-1\},
\label{eq:sample_times}
\end{equation}
ensuring accurate physical simulation of the exposure process while maintaining computational efficiency.
%%%%%%%%%%%%%%%%%%%%%%%%%%%%

\noindent\textbf{Blur Aware Tracking Optimization}
\label{sec:track_opt}

The tracking front-end is the workhorse of our SLAM system, responsible for estimating the camera's precise motion for each incoming blurry frame. Its goal is to find the most probable continuous trajectory, parameterized by the exposure start and end poses $(\mat{T}_{st}, \mat{T}_{ed})$. To achieve this, for each incoming blurry frame $m$, we estimate the exposure endpoints $(\mat{T}_{st}, \mat{T}_{ed})$ by minimizing
\begin{equation}
    \min_{\mat{T}_{st}, \mat{T}_{ed}} \mathcal{L}_{\text{photo}} + \lambda_{\text{evt}}\mathcal{L}_{\text{LI}},
    \label{eq:track_obj}
\end{equation}
where $\lambda_{\text{evt}}$ balances the event terms.

\textbf{1. Photometric Loss.} The foundational constraint of our optimization is photometric consistency. This principle dictates that our model encompassing both the 3D scene representation and the estimated camera trajectory must accurately replicate the physical image formation process. Consequently, the blurred image synthesized by our renderer should closely align with the actual image captured by the camera sensor. To enforce this, we synthesize a blurred image $\bar{\mat{C}}_m$ using our physics-based renderer, complete with the event-modulated spatial and temporal kernels. We then define the photometric loss as a robust error metric (e.g., Charbonnier loss $\rho$) between our rendered output and the observation $\mat{I}_{\text{obs}}$:
\begin{equation}
    \mathcal{L}_{\text{photo}} = \|\bar{\mat{C}}_m - \mat{I}_{\text{obs}}\|_\rho,
\end{equation}
where $\bar{\mat{C}}_m$ is rendered with per-Gaussian adaptive temporal samples $n_j^{(m)}$ and spatial modulation $\mat{\Sigma}^{2D}_{j,\text{prior}}$.

\textbf{2. Event Alignment Loss.} The event camera tells us exactly where and when brightness changed. Our estimated camera motion must be able to reproduce these same brightness changes in the rendered 3D scene. This loss directly aligns the rendered intensity changes with the observed events. Following the event generation model, an event's polarity $p_i \in \{+1, -1\}$ should correspond to the sign of the logarithmic intensity change. We penalize any deviation from this model:
\begin{equation}
    \mathcal{L}_{\text{LI}} = \sum_{e_i \in \mathcal{E}} \rho\!\left(p_i - \frac{\log I(t_i) - \log I(t_i - \delta)}{\theta}\right),
\end{equation}
where $I(t)$ is the rendered intensity at an event's timestamp $t_i$, $\delta$ is a small temporal offset, and $\theta$ is the event camera's contrast threshold. This provides strong geometric constraints, especially in visually ambiguous or low-texture regions.

\subsection{Mapping with Keyframe Management}
\label{sec:mapping}
While the tracker estimates frame-to-frame motion, the mapping back-end ensures long-term map consistency. It operates on a sliding window of keyframes to jointly refine 3D Gaussians and camera poses, maintaining physical agreement between all sensor observations over time.
\subsubsection{Keyframe Selection}
% 使用标准 LaTeX 三连字符 em-dash
We insert a new keyframe when either: (1) the view overlap with the latest 
keyframe measured as IoU of visible Gaussians drops below $\tau_{\text{overlap}}$, or (2) camera translation exceeds $\tau_{\text{trans}}\cdot \bar d$ (where $\bar d$ is average scene depth). When the keyframe limit is reached, we remove the keyframe with highest neighbor overlap to preserve view diversity while bounding computation.
% ======================== Rendering table ========================
\renewcommand{\arraystretch}{1.1}
\begin{table*}[htbp]
\caption{\textbf{Rendering performance comparison on \emph{EventReplica}}.
``--'' indicates sequences where reinitialization did not succeed.
Boldface marks the best result per column.}
\centering
 \resizebox{0.92\linewidth}{!}{
\begin{tabular}{cccccccccc||c}
\toprule
\textbf{Method} & \textbf{Metric} & \textbf{room0} & \textbf{room1} & \textbf{room2} & \textbf{office0} & \textbf{office1} & \textbf{office3} & \textbf{office4} & \textbf{Avg.} & \textbf{Rendering FPS} \\
\midrule
\multirow{3}{*}{PhotoSLAM~\cite{huang2024photo}} 
& PSNR[dB]$\uparrow$  & 17.91 & 18.96 & -- & 23.15 & -- & 17.04 & 14.28 & -- & \multirow{3}{*}{1058.6}\\
& SSIM$\uparrow$      & 0.499 & 0.588 & -- & 0.640 & -- & 0.596 & 0.546 & -- & \\
& LPIPS$\downarrow$   & 0.460 & 0.455 & -- & 0.417 & -- & 0.391 & 0.526 & -- & \\ 
\hline
\multirow{3}{*}{MonoGS~\cite{monogs}} 
& PSNR[dB]$\uparrow$  & 20.27 & 21.77 & 23.85 & 26.55 & 26.56 & 21.21 & 21.59 & 23.12 & \multirow{3}{*}{1102.1}\\
& SSIM$\uparrow$      & 0.612 & 0.677 & 0.748 & 0.769 & 0.810 & 0.729 & 0.749 & 0.728 & \\
& LPIPS$\downarrow$   & 0.474 & 0.471 & 0.355 & 0.385 & 0.355 & 0.287 & 0.411 & 0.391 & \\
\hline
\multirow{3}{*}{EGS-SLAM~\cite{chen2025egs}} 
& PSNR[dB]$\uparrow$  & 24.06 & 26.30 & 27.61 & 31.72 & 33.38 & 26.50 & 23.79 & 27.62 & \multirow{3}{*}{1168.3}\\
& SSIM $\uparrow$     & 0.744 & 0.783 & 0.838 & 0.885 & 0.927 & 0.846 & 0.806 & 0.833 & \\
& LPIPS$\downarrow$   & 0.229 & 0.256 & 0.172 & 0.142 & 0.123 & 0.113 & 0.242 & 0.182 & \\
\hline
\multirow{3}{*}{\textbf{MotionGS-SLAM (Ours)}} 
& PSNR[dB]$\uparrow$  & \textbf{24.55} & \textbf{27.05} & \textbf{28.22} & \textbf{32.31} & \textbf{33.92} & \textbf{26.98} & \textbf{24.10} & \textbf{28.05} & \multirow{3}{*}{\textbf{1189.7}}\\
& SSIM $\uparrow$     & \textbf{0.756} & \textbf{0.796} & \textbf{0.848} & \textbf{0.893} & \textbf{0.933} & \textbf{0.855} & \textbf{0.814} & \textbf{0.842} & \\
& LPIPS$\downarrow$   & \textbf{0.214} & \textbf{0.241} & \textbf{0.159} & \textbf{0.131} & \textbf{0.116} & \textbf{0.107} & \textbf{0.232} & \textbf{0.171} & \\
\bottomrule
\end{tabular}
}

\label{table:ReplicaRender}
\vspace{-0.5em}
\end{table*}
\subsubsection{Bundle Adjustment}
The core of our mapping process is a bundle adjustment over the sliding window $\mathcal{W}$, where we jointly optimize the Gaussian map $\mathcal{G}$ and the keyframe poses $\{\mat T\}$ using a comprehensive, blur-consistent objective:

\begin{equation}
\begin{aligned}
    \min_{\mathcal{G}, \{\mat{T}\}}~ 
    &\sum_{w \in \mathcal{W}} \Big[
        \mathcal{L}_{\text{photo}}^w 
        + \lambda_{\text{evt}}\mathcal{L}_{\text{LI}}^w
    \Big] \\
    &\qquad + \lambda_{\text{shape}} \mathcal{L}_{\text{shape}},
\end{aligned}
\label{eq:map_obj}
\end{equation}
where $\mathcal{W}$ is the set of keyframes in the current optimization window. This objective has two main parts:
\textbf{Data Consistency Terms:} The first summation, $\sum_{w \in \mathcal{W}}[\dots]$, ensures that for every keyframe in the window, our optimized map and poses can accurately reproduce the observed blurry images and event streams. This enforces consistency across time.

% Bundle Adjustment 段落中，将原来的 Eq.(15) 整块替换为：
\textbf{$\mathcal{L}_{\text{shape}}$: Image-Plane Shape Prior Loss.}
This loss encourages each 3D Gaussian, after projection onto the image 
plane, to form a 2D elliptical kernel slightly elongated along the detected 
motion direction, without altering its underlying 3D geometry. 
It is defined in Eq.~\eqref{eq:lshape}: if events detect motion in a 
certain direction, the projected Gaussian ellipse is encouraged to be 
gently elongated along that direction; otherwise, it remains isotropic.
This enforces a physically meaningful explanation of blur while avoiding 
contaminating the underlying 3D geometry.
% ← 完全删除原来的 equation 环境和第二个 \label{eq:lshape}

\section{Experiments}

In this section, we conduct a series of rigorous experiments to validate the effectiveness of MotionGS-SLAM. 

\subsection{Datasets and Metrics}

\textbf{Datasets:} We evaluate our method on both synthetic and real-world sequences to ensure a thorough analysis.
\par \textbf{EventReplica:} We created a challenging synthetic dataset by adapting the Replica dataset~\cite{straub2019replica}. For each camera trajectory, we render a high number of intermediate frames and simulate a long exposure by averaging them, which produces significant and physically realistic motion blur. To simulate low-light-induced motion blur conditions, we also inject Poisson-Gaussian noise into the final blurred images. This dataset provides ground truth sharp images for objective evaluation of reconstruction quality.
    Event streams are synthesized via ESIM~\cite{rebecq2018esim} using a balanced threshold $\Theta=0.2$ and monochrome events.
\par \textbf{Real Database:} To test our system in real-world scenarios, we introduce a real-world dataset of $3$ indoor/outdoor scenes recorded with a Color-DAVIS346~\cite{taverni2018front} camera. This device captures both color frames at $346 \times 260$ resolution and color events. We capture 3 challenging scenes using a handheld camera setup under low-light conditions. Each scene contains rich textures and color information. For each scene, we acquire 30s frames across multiple viewpoints with varying degrees of motion blur, along with temporally aligned event streams recorded by the event camera.

\textbf{Metrics:} We use standard metrics for evaluation. For trajectory accuracy, we report the \textbf{Absolute Trajectory Error (ATE)} [cm]. For mapping quality on the synthetic dataset, we report \textbf{PSNR}, \textbf{SSIM}, and \textbf{LPIPS}~\cite{zhang2018unreasonable}. 

\subsection{Implementation Details}

Our system is implemented in PyTorch and runs on a single NVIDIA RTX 4080 GPU. We build our system upon the MonoGS~\cite{monogs} codebase. For our event-modulated kernel, we set the base temporal samples $n_0=9$ and the spatial stretch factor control $\beta=0.05$. In our loss function, weights are set to $\lambda_\text{evt}=2.0$, $\lambda_\text{photo}=1.0$, and $\lambda_\text{shape}=0.2$. For all comparisons, we ensure that competing methods are configured to their official recommended settings for a fair evaluation. In all tables, \textbf{boldface} indicates the best result.

\subsection{Comparison with State-of-the-Art}

\noindent\textbf{Evaluation on Blurry-Replica with Event (Synthetic)}
We compare \textbf{MotionGS-SLAM} against representative baselines on the challenging \emph{EventReplica} benchmark under low-light and long-exposure settings: the photometric SLAM \textbf{PhotoSLAM}~\cite{huang2024photo}, the image-only GS-SLAM \textbf{MonoGS}~\cite{monogs}, and the event-guided \textbf{EGS-SLAM}. We evaluate two aspects: (i) reconstruction fidelity, and (ii) tracking accuracy under severe motion blur.

\noindent\textbf{Reconstruction quality and throughput.}
Table~\ref{table:ReplicaRender} reports PSNR/SSIM/LPIPS and rendering FPS. While PhotoSLAM and MonoGS degrade notably with persistent blur, and EGS-SLAM improves upon image-only pipelines by utilizing events, \textbf{MotionGS-SLAM} achieves the best fidelity across almost all sequences and the highest rendering FPS. Concretely, \emph{MotionGS-SLAM} improves the \emph{average} PSNR from \emph{23.12\,dB} (MonoGS) and \emph{27.62\,dB} (EGS-SLAM) to \textbf{28.05\,dB}; SSIM from \emph{0.728} (MonoGS) and \emph{0.833} (EGS-SLAM) to \textbf{0.842}; and LPIPS from \emph{0.391} (MonoGS) and \emph{0.182} (EGS-SLAM) down to \textbf{0.171}. On representative sequences, our PSNR is higher (e.g., \texttt{office1}: \textbf{33.92} vs.\ 33.38; \texttt{office0}: \textbf{32.31} vs.\ 31.72; \texttt{room1}: \textbf{27.05} vs.\ 26.30). Throughput-wise, \emph{MotionGS-SLAM} reaches \textbf{1189.7} FPS, exceeding EGS-SLAM (1168.3), MonoGS (1102.1), and PhotoSLAM (1058.6), indicating that event-modulated spatial/temporal kernels not only sharpen reconstructions but also preserve rendering speed.
\begin{figure}[tbp]
    \centering
    \includegraphics[width=1\linewidth]{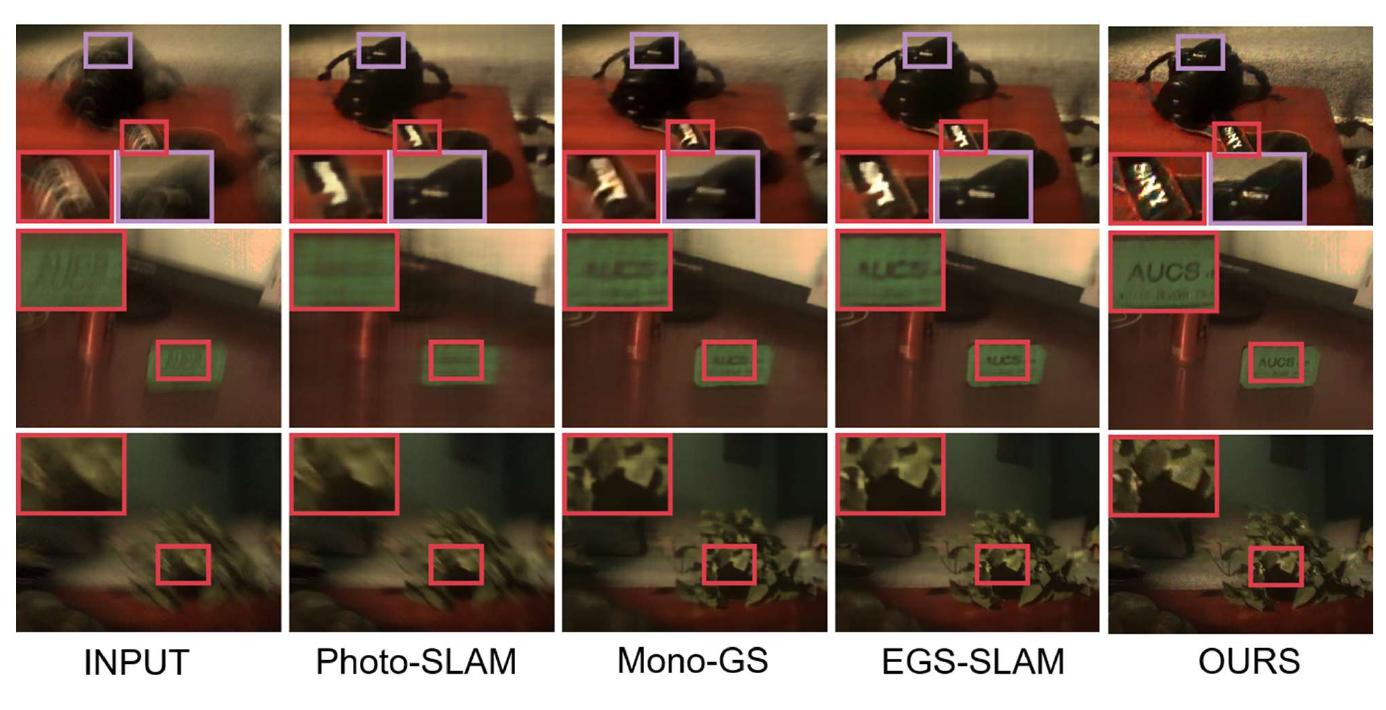} % 
    \caption{\noindent\textbf{Qualitative comparison on our real-world dataset.} We demonstrate our method's performance on challenging, low-light handheld sequences. As shown in the corresponding figure, image-only baselines like Photo-SLAM~\cite{huang2024photo} and Mono-GS~\cite{monogs} produce blurry reconstructions with severe "ghosting" artifacts. While the event-based EGS-SLAM~\cite{chen2025egs} improves upon them, it still suffers from soft details. In contrast, our method is the only one that successfully reconstructs a sharp, geometrically consistent scene, recovering fine details.}
    \label{fig:comparison_real}
    \vspace{-8pt}
\end{figure}
\noindent\textbf{Tracking accuracy.}
As shown in Table~\ref{table:eventreplica_ate}, \textbf{MotionGS-SLAM} achieves the lowest ATE in all seven sequences. ORB-SLAM2 and PhotoSLAM frequently lose tracking as blurred edges hamper keypoint detection; MonoGS assumes blur-free inputs and suffers on long-exposure frames. EGS-SLAM benefits from events but lacks our exposure-integration modeling and per-Gaussian modulation, leading to consistently higher errors. Overall, \emph{MotionGS-SLAM} reduces the average ATE from \emph{9.22}\,cm (MonoGS) to \textbf{4.44}\,cm ($\sim$51.9\% improvement), and further improves over EGS-SLAM (5.17\,cm) by $\sim$14.2\%. These numbers exactly correspond to the averages reported in Table~\ref{table:eventreplica_ate}.

\noindent\textbf{Evaluation on Real-World Data}
We further validate our method on three challenging, low-light handheld scenes. As shown in Table~\ref{tab:real_world_combined}, the event-based EGS-SLAM is the strongest baseline, significantly outperforming image-only methods. 

% ======================== NEW COMBINED REAL DATA TABLE (ATE ONLY) V2 ========================
\begin{table}[htbp]
\centering
\caption{\textbf{Tracking Accuracy (ATE in cm) and Ablation Study on our Real-World Dataset}. }
\label{tab:real_world_combined}
\resizebox{\columnwidth}{!}{%
\begin{tabular}{lcccc}
\toprule
\textbf{Method} & \textbf{Scene 1} & \textbf{Scene 2} & \textbf{Scene 3} & \textbf{Avg.} \\
\midrule
\multicolumn{5}{l}{\textit{Comparison with State-of-the-Art}} \\
ORB-SLAM3~\cite{orbslam3} & 8.91 & 35.42 & 41.05 & 28.46 \\
PhotoSLAM~\cite{huang2024photo} & 11.24 & 9.85 & 28.66 & 16.58 \\
EGS-SLAM ~\cite{chen2025egs}& 4.52 & 5.15 & 4.81 & 4.83 \\
\midrule
\multicolumn{5}{l}{\textit{Ablation Study}} \\
Baseline (MonoGS)~\cite{monogs} & 6.88 & 7.12 & 5.95 & 6.65 \\
+ Events & 4.65 & 5.01 & 4.42 & 4.69 \\
+ Spatial Modulation & 4.03 & 4.45 & 3.88 & 4.12 \\
\textbf{Full Model (Ours)} & \textbf{3.45} & \textbf{4.02} & \textbf{3.18} & \textbf{3.55} \\
\bottomrule
\end{tabular}%
}
\end{table}

% ======================== Tracking table ========================
\begin{table}[htbp]
\centering
\caption{\textbf{Tracking results ATE (cm) on \emph{EventReplica}}. \textcolor{red}{\textbf{L}} denotes tracking failure. Boldface marks the best result per row/average.}
\label{table:eventreplica_ate}
\resizebox{0.98\linewidth}{!}{
\begin{tabular}{lccccc}
\toprule
\textbf{Scenes} & \makecell[c]{ORB-SLAM2\\~\cite{orbslam2}} & \makecell[c]{PhotoSLAM\\~\cite{huang2024photo}} & \makecell[c]{MonoGS\\~\cite{monogs}} & \makecell[c]{EGS-SLAM\\~\cite{chen2025egs}} & \textbf{MotionGS-SLAM} \\
\midrule
\textbf{room0}    & 5.57 & 6.64 & 12.76 & 4.85 & \textbf{3.98} \\
\textbf{room1}    & \textcolor{red}{\textbf{L}} & \textcolor{red}{\textbf{L}} & 8.45 & 3.55 & \textbf{2.88} \\
\textbf{room2}    & \textcolor{red}{\textbf{L}} & \textcolor{red}{\textbf{L}} & 3.64 & 3.25 & \textbf{2.84} \\
\textbf{office0}  & \textcolor{red}{\textbf{L}} & \textcolor{red}{\textbf{L}} & 7.44 & 3.75 & \textbf{3.01} \\
\textbf{office1}  & \textcolor{red}{\textbf{L}} & \textcolor{red}{\textbf{L}} & 7.78 & 3.66 & \textbf{3.06} \\
\textbf{office3}  & 6.48 & 6.78 & 7.91 & 4.55 & \textbf{3.92} \\
\textbf{office4}  & \textcolor{red}{\textbf{L}} & \textcolor{red}{\textbf{L}} & 16.55 & 12.60 & \textbf{11.37} \\
\hline
\textbf{Avg.}     & --   & --   & 9.22 & 5.17 & \textbf{4.44} \\
\bottomrule
\vspace{-5pt}
\end{tabular}
}
\end{table}

\textbf{Qualitative Results:}
Fig.3 and Fig.4 provide a visual comparison of the reconstructed maps. The outputs from MonoGS are visibly blurry and contain significant floating artifacts and "ghosting" effects, consistent with the quantitative results. In contrast, our reconstruction is sharp, detailed, and geometrically coherent, highlighting the power of our physics-based, event-guided rendering approach.

\subsection{Ablation Study}
\label{sec:ablation}

To rigorously validate our design choices, we conduct a detailed ablation study on two representative \emph{EventReplica} sequences. We analyze how each component of our event-guided framework contributes to the final performance. The results are summarized in Table~\ref{tab:ablation_eventreplica} and shown in Fig. 5.

Our ablation compares five system variants:
\textbf{A0 (Baseline):} A standard GS-SLAM system (MonoGS) using only the blurry images, with no event data or explicit blur model. \textbf{A1 (Blur Model Only):} Augments the baseline with our physics-based exposure integration renderer, but without any guidance from events. \textbf{A2 (Event Assoc.):} Introduces event data and association but does not yet use it to modulate the rendering kernel. Events contribute only through standard losses ($\mathcal{L}_{\text{LI}}$. \textbf{A3 (Spatial Modulation Only):} Builds on A2 by adding our spatial kernel modulation (anisotropic deformation via $\mathcal{L}_{\text{shape}}$), but keeps temporal sampling fixed. \textbf{A4 (Temporal Modulation Only):} Builds on A2 by adding our adaptive temporal sampling, but keeps the spatial kernel untouched. \textbf{A5 (Full / Ours):} The complete MotionGS-SLAM system with both spatial and temporal modulation enabled.

\renewcommand{\arraystretch}{1.1}
\begin{table}[htbp]
\centering
\caption{\textbf{Ablation on \emph{EventReplica}} for two scenes. Best results in each column are in \textbf{bold}.}
\label{tab:ablation_eventreplica}
\resizebox{0.98\linewidth}{!}{
\begin{tabular}{l|cccc|cccc}
\toprule
\multirow{2}{*}{Variant} & \multicolumn{4}{c|}{\texttt{room0}} & \multicolumn{4}{c}{\texttt{office0}} \\
& ATE$\downarrow$ & PSNR$\uparrow$ & SSIM$\uparrow$ & LPIPS$\downarrow$ & ATE$\downarrow$ & PSNR$\uparrow$ & SSIM$\uparrow$ & LPIPS$\downarrow$ \\
\midrule
A0\_Baseline        & 12.76 & 20.27 & 0.612 & 0.474 & 7.44 & 26.55 & 0.769 & 0.385 \\
A1\_Blur\_Only      & 11.80 & 21.10 & 0.630 & 0.455 & 6.90 & 27.30 & 0.780 & 0.370 \\
A2\_Event\_Assoc    &  6.10 & 22.20 & 0.700 & 0.340 & 4.10 & 28.50 & 0.830 & 0.240 \\
A3\_Spatial\_Only   &  5.40 & 23.60 & 0.735 & 0.250 & 3.50 & 30.70 & 0.880 & 0.160 \\
A4\_Temporal\_Only  &  5.10 & 23.10 & 0.725 & 0.265 & 3.40 & 30.20 & 0.870 & 0.175 \\
A5\_Full (Ours)     & \textbf{3.98} & \textbf{24.55} & \textbf{0.756} & \textbf{0.214} & \textbf{3.01} & \textbf{32.31} & \textbf{0.893} & \textbf{0.131} \\
\bottomrule
\end{tabular}
}
\end{table}
\paragraph{The Necessity of Event Data}
The results in Table~\ref{tab:ablation_eventreplica} reveal the critical role of event cameras in addressing motion blur. The baseline (A0) fails catastrophically with blur-only inputs, while simply adding our physics-based blur model (A1) provides marginal 
gains reducing ATE from 12.76cm to 11.80cm on \texttt{room0}. The breakthrough comes with event integration (A2): ATE plummets to 6.10cm, a 52\% reduction from baseline. This dramatic improvement demonstrates that events provide the missing temporal resolution needed to disambiguate motion during exposure. However, A2's relatively poor perceptual metrics (LPIPS of 0.340 vs. our full model's 0.214) exposes a crucial limitation: standard event losses alone cannot solve the rendering problem. They constrain camera motion but fail to inform how blur should be synthesized.

\paragraph{Dissecting the Dual-Modulation Kernel}
Comparing variants A2 through A5 validates our core hypothesis: effective blur handling requires modeling both its spatial formation and temporal integration. \textbf{Spatial modulation (A3)} transforms the reconstruction quality LPIPS improves from 0.340 to 0.250 on \texttt{room0}, the single largest perceptual gain across all components. This confirms our insight that event-guided anisotropic deformation correctly models the physical "painting" of motion streaks. \textbf{Temporal modulation (A4)} contributes differently but equally importantly: it ensures photometric accuracy by adapting the exposure integral's sampling density to local motion, improving PSNR while maintaining computational efficiency. The \textbf{full model (A5)} demonstrates clear synergy: combining both modulations yields the best performance across all metrics (ATE: 3.98cm, PSNR: 24.55dB, LPIPS: 0.214), surpassing either component alone. This validates our paradigm shift rather than attempting to invert blur, we must model both \textit{how blur spatially manifests} and \textit{how it temporally accumulates} to achieve robust SLAM under severe motion.

\begin{figure}[tbp]
    \centering
    \includegraphics[width=0.97\linewidth]{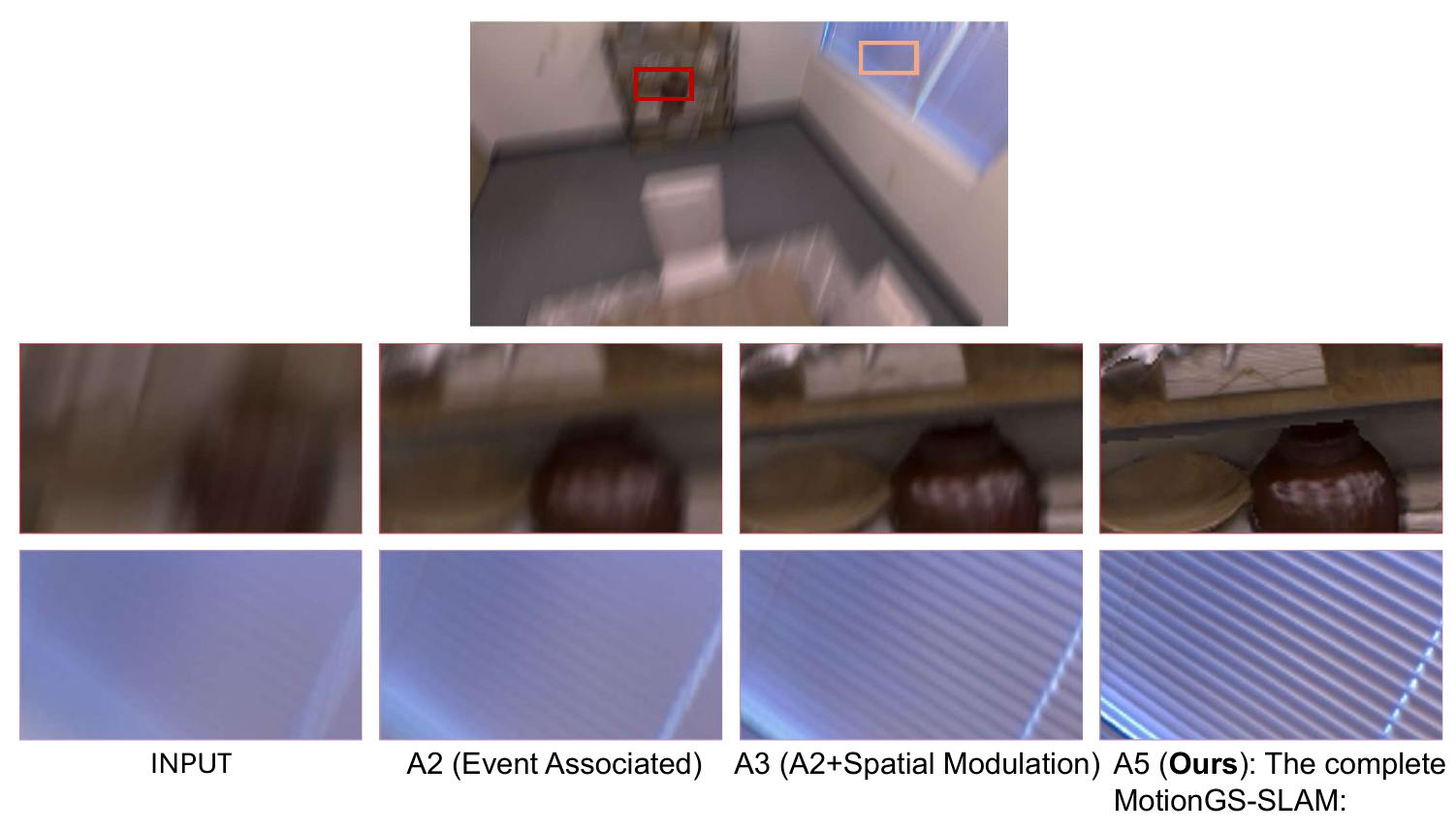} % 
    \vspace{-5pt}\caption{\textbf{Qualitative Ablation Study on EventReplica Dataset.} Top row shows severely blurred input. Red and orange boxes highlight reconstructed details for: \textbf{A2:} Event data with standard losses. \textbf{A3:} A2 + spatial kernel modulation. \textbf{A5 (Ours):} Complete MotionGS-SLAM. Clear progression from blurry A2 to sharp A3, culminating in high-fidelity reconstruction with our full model.}
    \vspace{-5pt}\label{fig:comparison_albation}
\end{figure}

\section{Summary}

We introduce \textbf{MotionGS-SLAM}, a robust SLAM system that excels in high-motion scenarios by fundamentally shifting the approach to motion blur. Instead of treating blur as an artifact to be corrected, we reformulate it as a physical process to be generatively modeled within the rendering pipeline. Our central contribution is a novel \textbf{event-modulated Gaussian kernel}, which leverages high-frequency event data to guide this physics-based synthesis. By transforming blur from a challenge into a supervisory signal, our method achieves state-of-the-art tracking accuracy and sharp scene reconstruction directly from severely degraded images.

%%%%%%%%%%%%%%%%%%%%%%%%%%%%%%%
%\thispagestyle{empty}
	{\small
	\bibliographystyle{ieee_fullname}
	\bibliography{egbib}}

@article{orbslam2,
  title={Orb-slam2: An open-source slam system for monocular, stereo, and rgb-d cameras},
  author={Mur-Artal, Raul and Tard{\'o}s, Juan D},
  journal={IEEE Trans. Robot.},
  volume={33},
  number={5},
  pages={1255--1262},
  year={2017}
}

@article{orbslam3,
  title={Orb-slam3: An accurate open-source library for visual, visual--inertial, and multimap slam},
  author={Campos, Carlos and Elvira, Richard and Rodr{\'\i}guez, Juan J G and others},
  journal={IEEE Trans. Robot.},
  volume={37},
  number={6},
  pages={1874--1890},
  year={2021}
}

@article{droidslam,
  title={Droid-slam: Deep visual slam for monocular, stereo, and rgb-d cameras},
  author={Teed, Zachary and Deng, Jia},
  journal={Advances in neural information processing systems},
  volume={34},
  pages={16558--16569},
  year={2021}
}

@inproceedings{monogs,
  title={Gaussian splatting slam},
  author={Matsuki, Hidenobu and Murai, Riku and Kelly, Paul HJ and Davison, Andrew J},
  booktitle={Proceedings of the IEEE/CVF Conference on Computer Vision and Pattern Recognition},
  pages={18039--18048},
  year={2024}
}

@inproceedings{gsslam,
  title={Gs-slam: Dense visual slam with 3d gaussian splatting},
  author={Yan, Chi and Qu, Delin and Xu, Dan and Zhao, Bin and Wang, Zhigang and Wang, Dong and Li, Xuelong},
  booktitle={Proceedings of the IEEE/CVF Conference on Computer Vision and Pattern Recognition},
  pages={19595--19604},
  year={2024}
}

@inproceedings{splatam,
  title={Splatam: Splat track \& map 3d gaussians for dense rgb-d slam},
  author={Keetha, Nikhil and Karhade, Jay and Jatavallabhula, Krishna Murthy and Yang, Gengshan and Scherer, Sebastian and Ramanan, Deva and Luiten, Jonathon},
  booktitle={Proceedings of the IEEE/CVF Conference on Computer Vision and Pattern Recognition},
  pages={21357--21366},
  year={2024}
}

@inproceedings{huang2024photo,
  title={Photo-slam: Real-time simultaneous localization and photorealistic mapping for monocular stereo and rgb-d cameras},
  author={Huang, Huajian and Li, Longwei and Cheng, Hui and Yeung, Sai-Kit},
  booktitle={Proceedings of the IEEE/CVF Conference on Computer Vision and Pattern Recognition},
  pages={21584--21593},
  year={2024}
}

@inproceedings{i2slam,
  title={I 2-SLAM: Inverting Imaging Process for Robust Photorealistic Dense SLAM},
  author={Bae, Gwangtak and Choi, Changwoon and Heo, Hyeongjun and Kim, Sang Min and Kim, Young Min},
  booktitle={European Conference on Computer Vision},
  pages={72--89},
  year={2024},
  organization={Springer}
}

@article{xiong2024event3dgs,
  title={Event3dgs: Event-based 3d gaussian splatting for high-speed robot egomotion},
  author={Xiong, Tianyi and Wu, Jiayi and He, Botao and Fermuller, Cornelia and Aloimonos, Yiannis and Huang, Heng and Metzler, Christopher A},
  journal={arXiv preprint arXiv:2406.02972},
  year={2024}
}

@inproceedings{deblurnerf,
  title={Deblur-nerf: Neural radiance fields from blurry images},
  author={Ma, Li and Li, Xiaoyu and Liao, Jing and others},
  booktitle={Proc. IEEE/CVF Conf. Comput. Vis. Pattern Recognit.},
  pages={12861--12870},
  year={2022}
}

@inproceedings{qi2023e2nerf,
  title={E2nerf: Event enhanced neural radiance fields from blurry images},
  author={Qi, Yunshan and Zhu, Lin and Zhang, Yu and Li, Jia},
  booktitle={Proceedings of the IEEE/CVF International Conference on Computer Vision},
  pages={13254--13264},
  year={2023}
}

@inproceedings{eventnerfslam,
  title={Implicit event-rgbd neural slam},
  author={Qu, Delin and Yan, Chi and Wang, Dong and Yin, Jige and others},
  booktitle={Proc. IEEE/CVF Conf. Comput. Vis. Pattern Recognit.},
  pages={19584--19594},
  year={2024}
}

@article{gs,
  title={3D Gaussian splatting for real-time radiance field rendering.},
  author={Kerbl, Bernhard and Kopanas, Georgios and Leimk{\"u}hler, Thomas and Drettakis, George},
  journal={ACM Trans. Graph.},
  volume={42},
  number={4},
  pages={139--1},
  year={2023}
}

@inproceedings{yu2024evagaussians,
  title={Evagaussians: Event stream assisted gaussian splatting from blurry images},
  author={Yu, Wangbo and Feng, Chaoran and Li, Jianing and Tang, Jiye and Yang, Jiashu and Tang, Zhenyu and Cao, Meng and Jia, Xu and Yang, Yuchao and Yuan, Li and others},
  booktitle={Proceedings of the IEEE/CVF International Conference on Computer Vision},
  pages={24780--24790},
  year={2025}
}

@article{han2024event,
  title={Event-3dgs: Event-based 3d reconstruction using 3d gaussian splatting},
  author={Han, Hanqian and Li, Jianing and Wei, Henglu and Ji, Xiangyang},
  journal={Advances in Neural Information Processing Systems},
  volume={37},
  pages={128139--128159},
  year={2024}
}

@inproceedings{rebecq2018esim,
  title={Esim: an open event camera simulator},
  author={Rebecq, Henri and Gehrig, Daniel and Scaramuzza, Davide},
  booktitle={Conference on robot learning},
  pages={969--982},
  year={2018},
  organization={PMLR}
}

@article{taverni2018front,
  title={Front and back illuminated dynamic and active pixel vision sensors comparison},
  author={Taverni, Gemma and Moeys, Diederik Paul and Li, Chenghan and Cavaco, Celso and Motsnyi, Vasyl and Bello, David San Segundo and Delbruck, Tobi},
  journal={IEEE Transactions on Circuits and Systems II: Express Briefs},
  volume={65},
  number={5},
  pages={677--681},
  year={2018},
  publisher={IEEE}
}

@article{chen2025egs,
  title={EGS-SLAM: RGB-D Gaussian Splatting SLAM with Events},
  author={Chen, Siyu and Yuan, Shenghai and Nguyen, Thien-Minh and Huang, Zhuyu and Shi, Chenyang and Jing, Jin and Xie, Lihua},
  journal={IEEE Robotics and Automation Letters},
  year={2025},
  publisher={IEEE}
}

@inproceedings{li2024sgs,
  title={Sgs-slam: Semantic gaussian splatting for neural dense slam},
  author={Li, Mingrui and Liu, Shuhong and Zhou, Heng and Zhu, Guohao and Cheng, Na and Deng, Tianchen and Wang, Hongyu},
  booktitle={European Conference on Computer Vision},
  pages={163--179},
  year={2024},
  organization={Springer}
}

@article{fan2024large,
  title={Large spatial model: End-to-end unposed images to semantic 3d},
  author={Fan, Zhiwen and Zhang, Jian and Cong, Wenyan and Wang, Peihao and Li, Renjie and Wen, Kairun and Zhou, Shijie and Kadambi, Achuta and Wang, Zhangyang and Xu, Danfei and others},
  journal={Advances in neural information processing systems},
  volume={37},
  pages={40212--40229},
  year={2024}
}

@article{deng2024compact,
  title={Compact 3d gaussian splatting for dense visual slam},
  author={Deng, Tianchen and Chen, Yaohui and Zhang, Leyan and Yang, Jianfei and Yuan, Shenghai and Liu, Jiuming and Wang, Danwei and Wang, Hesheng and Chen, Weidong},
  journal={arXiv preprint arXiv:2403.11247},
  year={2024}
}

@inproceedings{cannici2024mitigating,
  title={Mitigating motion blur in neural radiance fields with events and frames},
  author={Cannici, Marco and Scaramuzza, Davide},
  booktitle={Proceedings of the IEEE/CVF Conference on Computer Vision and Pattern Recognition},
  pages={9286--9296},
  year={2024}
}

@inproceedings{zhang2018unreasonable,
  title={The unreasonable effectiveness of deep features as a perceptual metric},
  author={Zhang, Richard and Isola, Phillip and Efros, Alexei A and Shechtman, Eli and Wang, Oliver},
  booktitle={Proceedings of the IEEE conference on computer vision and pattern recognition},
  pages={586--595},
  year={2018}
}

@article{straub2019replica,
  title={The replica dataset: A digital replica of indoor spaces},
  author={Straub, Julian and Whelan, Thomas and Ma, Lingni and Chen, Yufan and Wijmans, Erik and Green, Simon and Engel, Jakob J and Mur-Artal, Raul and Ren, Carl and Verma, Shobhit and others},
  journal={arXiv preprint arXiv:1906.05797},
  year={2019}
}

%\end{thebibliography}

\end{document}